\documentclass[sigconf]{acmart}
\usepackage[utf8]{inputenc}
\usepackage{amsthm}
\usepackage{hyperref}
\usepackage{multirow}
\usepackage{booktabs}
\usepackage{algpseudocode}

\usepackage{amsmath,amsfonts,bm}

\def\eqref#1{equation~\ref{#1}}

\def\1{\bm{1}}

\DeclareMathAlphabet{\mathsfit}{\encodingdefault}{\sfdefault}{m}{sl}
\SetMathAlphabet{\mathsfit}{bold}{\encodingdefault}{\sfdefault}{bx}{n}

\newcommand{\jn}[1]{}
\renewcommand{\jn}[1]{{\color{red} JN: {#1}}}

\newcommand{\mcrot}[4]{\multicolumn{#1}{#2}{\rlap{\rotatebox{#3}{#4}~}}}

\title{Creating generalizable downstream graph models with random projections}
\author{Anton Amirov}
\email{antonam@microsoft.com}
\affiliation{
  \institution{Microsoft Corporation}
}
\author{Chris Quirk}
\email{chrisq@microsoft.com}
\affiliation{
  \institution{Microsoft Corporation}
}
\author{Jennifer Neville}
\email{jenneville@microsoft.com}
\affiliation{
  \institution{Microsoft Research}
}
\acmConference[MLoG Workshop at WSDM'23]{Machine Learning on Graphs Workshop at WSDM'23}{March 3, 2023}{Singapore}

\date{}

\begin{document}

\begin{abstract}

We investigate graph representation learning approaches that enable models to generalize across graphs: given a model trained using the representations from one graph, our goal is to apply inference using those same model parameters when given representations computed over a new graph, unseen during model training, with minimal degradation in inference accuracy. This is in contrast to the more common task of doing inference on the unseen nodes of the same graph. We show that using random projections to estimate multiple powers of the transition matrix allows us to build a set of isomorphism-invariant features that can be used by a variety of tasks. The resulting features can be used to recover enough information about the local neighborhood of a node to enable inference with relevance competitive to other approaches while maintaining computational efficiency.

\end{abstract}

\maketitle

\section{Introduction}

Graph-structured data is undergoing explosive growth, not only in large shared graphs (such as Pinterest, Facebook, and other social networks), but also in user and organization specific graphs -- stored in phones, corporate data centers, and isolated containers within the cloud.
Leveraging this graph-structured data to improve recommendation and prediction tasks can create substantial benefits for the user.
Vision and language tasks have been transformed by shared large-scale representation learning.
How do we build such shared representations for graphs, especially when those graphs are segmented or isolated? 
For instance, how can we build the embeddings of local businesses as nodes in a graph connecting businesses and users who review them in different cities, train a model that predicts some property of those businesses based on such embeddings using data from one city, and apply that model to another city?

We need to learn representations that generalize across graphs.
In this paper, we first formalize the notion of generalizable graph representation learning.
Furthermore, we present one such generalizable approach: a set of isomorphism-invariant features that capture rich information about the local neighborhood of the node.

In this work we look at the ways to build representations of nodes that incorporate information about the graph structure and the models that are using such representations to infer some properties of the nodes or pairs of nodes. Moreover, we want to create such representations in a task-agnostic way and in a way that generalizes prediction model to unseen graphs. Satisfying all three requirements (generalization, node level representation, agnostic to the final task) is important to understand motivation for the described approach, so we will give formal definition of each concept.

\subsection{Generalization}

Informally, we would like to learn representation methods that are portable across graphs and can be used for different tasks.
Consider natural language embedding methods like word2vec \cite{word2vec_mikolov}, GloVE \cite{glove_paper}, BERT \cite{bert_paper}: a single pre-trained model can be used on nearly any type of English text to produce a robust representation. As an example, a downstream model can rely on those representations to perform inference on words and phrases unseen during its training. 
However, constructing generic pre-trained models over multiple graphs is more challenging, because different graphs generally do not have a common vocabulary or foundation (in contrast to BERT models which can be applied to any English input).
Instead, each graph has its own inventory of nodes and edges -- almost like different languages, but often even more difficult to correlate, since different languages to a large degree operate over the similar concepts, hence the ability to translate from one language to another with minimal loss of meaning -- something that does not have a universal equivalent in the graph world.
Nevertheless, we would like to learn representations that port smoothly across a wide range of different graphs and practical applications.

This generalization property is defined over a distribution of graphs $\mathcal{G}$ and prediction task $\mathcal{T}$ in the following way.

\begin{itemize} 
\item 	Task $\mathcal{T}$ is a regression/classification task that maps an ordered set of nodes of fixed size to some label $Y$. In this work we are focusing on the most common types of tasks –- node level and pairwise predictions (set sizes 1 and 2 accordingly). Note that pairwise prediction is distinct from edge prediction, since it does not require that edge connection between the nodes exists in the graph.
\item   We have a set $N_G$ of training graphs and $N'_G$ of testing graphs, both drawn from the distribution $\mathcal{G}$.
\item   We have a (potentially partial) set of labels $Y_G$ for the task $\mathcal{T}$ over both the training and test graphs.
\item 	We train a model $\mathcal{M}$ on the training graphs $N_G$ and labels $Y_G$. Since we want the model to be applicable to any graph drawn from $\mathcal{G}$, it has to be inductive over the graph distribution and can be formulated as a function $\mathcal{M}(\Theta,\ V,\ E,\ i)$ for node or $\mathcal{M}(\Theta,\ V,\ E,\ i,\ j)$ for pairwise prediction, where $\Theta$ is the set of model parameters, $V$ – set of graph nodes (possibly with features), $E$ – set of graph edges, and $i$ and $j$ are the indexes of the nodes. Training on $N_G$  means that $\Theta$ is a function of $N_G$. Note that a model that is inductive across multiple graphs can be transductive within the graph and vice versa – those are independent properties. 
\item 	We also introduce a model fitness criteria $\mathcal{F}\left(\Theta,\ G,\ Y_G\right)$ that applies model $\mathcal{M}$ with parameters $\Theta$ to the graph $G$ to predict or assign probabilities to labels and compare these labels to actual labels $Y_G$, producing a numeric fitness score.
\item 	Now we can say that model $\mathcal{M}$ is generalizable on $\mathcal{G}$ if for the test sample $N_G^\prime$ drawn from the same distribution $\mathcal{G}$ the value of this fitness criteria is better than what can be obtained from the baseline model. The most trivial baseline is constant prediction optimizing the fitness criteria. From a practical standpoint, we often want to consider as baseline the model which does not use graph information (i.e., using only features of the node or pair of nodes for making predictions if such features exist). Note that this criterion is relatively weak, as we do not compare performance on the test set $N_G^\prime$ to performance on the training set $N_G$, nor do we consider nodes of graphs from $N_G$ whose labels were not available during training. The average gain in fitness criteria over a baseline on the test sample can be considered a measure of generalization that allows us to compare generalization capabilities of different models.
\end{itemize}

\subsection{Node level representation}
Models $\mathcal{M}(\Theta,\ V,\ E,\ i)$ and $\mathcal{M}(\Theta,\ V,\ E,\ i,\ j)$  from the definition above have unrestricted access to the information about the graph at inference time, for any given node or pair of nodes.
For instance, task specific 
graph neural networks typically use features of all the nodes in k-hop neighborhood of the node of interest. Without the use of caching, their run time is proportional to the average size of such a neighborhood and can be substantial in practical applications. Processing large neighborhoods is especially challenging when inference needs to happen in real-time. Caching GNN computations can improve preformance and also produces a node level representation. However, such representations will be task specific if the GNN is trained to solve some particular task. On the other hand, if a model relies solely on a node level representation, the model is forced to encode information about the node into the form of vector in some space  $\mathbb{R}^d$ by applying an embedding function that maps the graph into an embedding matrix with rows representing nodes
\[Q=\mathcal{E}\left(\mathrm{\Psi},\ V,\ E\right),\ Q\in\mathbb{R}^{|V|\times D}\]
and building models based on such embeddings (here and below $\Psi$ is used to refer to parameters of embedding model, and $\Theta$ to parameters of task specific model) 
\[\mathcal{M}\left(\Theta,\ V,\ E,\ i\right)=\mathcal{M}\left(\Theta,\ Q_{i,\ast}\right)\]
for single node regression and 
\[\mathcal{M}\left(\Theta,\ V,\ E,\ i,\ j\right)=\mathcal{M}\left(\Theta,\ Q_{i,\ast},\ Q_{j,\ast}\right)\]
for pairwise prediction.

\subsection{Task agnostic representations}
We want to create embeddings that work across multiple tasks, including those that are not known during the embedding creation process. Therefore, we want to avoid using task-specific labels Y to train the parameters $\mathrm{\Psi}$ of the embedding model. Such a restriction is not theoretical and can arise in practical settings for several reasons:

\begin{itemize} 
\item 	If model $\mathcal{E}$ is transductive within the graph, i.e., it contains node level parameters (trainable embeddings), training of such parameters relies on availability of task specific labels for each node. Such labels may be difficult to obtain (for instance, they may require expensive labeling or rely on telemetry that is not available for all nodes and/or becomes obsolete by the time it is collected and processed due to changes in the graph)
\item 	Even if model $\mathcal{E}$ is inductive, i.e. embeddings can be created for unlabeled nodes or for entirely new graphs, having task agnostic embeddings may be preferred if embeddings are shared between multiple downstream tasks which treat them as features that can be used to improve relevance. While in such setting, multi task training that jointly optimize model $\mathcal{E}$ with set of downstream tasks $\mathcal{M}_1,\ldots\mathcal{M}_K$ is likely to outperform task-agnostic embedding generation, it would require simultaneous training and update of parameters of all tasks $\Theta_1,\ldots\Theta_K$ after/with update of embedding model parameters $\mathrm{\Psi}$.  
Such joint training may be undesirable in business settings, since models $\mathcal{M}_1,\ldots\mathcal{M}_K$ may belong to different features, with separate shipping and training cadencies. Retraining of all models every time any of them iterates on its dataset or model structure can lead to unstable user experience and/or negatively affect development agility due to logistical challenges. 

\end{itemize}

\subsection{Problem formulation and summary}


Our goal is to create generalizable and task-agnostic node representations for graphs. These representations, similar to foundational NLP models, enable the construction of downstream models for multiple tasks that may not be known beforehand, and enables them to run on graphs that have not been seen during training time. Additionally, this allows the downstream model to bypass the need to traverse the graph, which makes inference time independent of the size of the graph or the node neighborhood.  

To achieve this goal, we will develop a method based on random projections. While there are existing approaches that can achieve the same goal, some of which we describe and compare against, we provide empirical evidence that the techniques we suggest can compete and in many cases provide relevance and/or computational gains over them. Our approaches are based on the ability of random projections to approximate multiple powers of the graph transition matrix in a computationally efficient way, and on the fact that such powers represent the properties of random walks of different lengths starting at a given node or pair of nodes, which are, in turn, isomorphic features capturing the properties of node neighborhoods of different sizes - from immediate neigborhood to the one spanning entire graph.


\section{Related work}

On a single graph, classification using task agnostic embeddings is a well explored problem. For instance, papers introducing random walk based techniques such as DeepWalk \cite{deepwalk_paper} and Node2Vec \cite{node2vec_paper} are evaluating performance on multi-label classification task. However, the way such embeddings are commonly used in single node inference tasks -- passing embedding vector to downstream model -- is not stable even against rotational symmetries (Node2Vec and DeepWalk models use vector dot product in the loss function, which is invariant against rotations of vector space, therefore embedding spaces resulting from different trainings on the same graph  at very minimum can be randomly rotated for different trainings aginst each other even on the same graph). While such embeddings can be used to create rotation invariant features, for many  single node classification/regression tasks the use of such features derived from single node embedding vector (or two vectors if we consider input and output embeddings) leads to weak downstream models with relevance barely above baseline. However, for pairwise classification, even simple dot product or euclidean distance between embeddings can lead to reasonable model quality, which can be further improved by some modification, as will be shown later.
In our experience, techniques like singular vector decomposition that create canonical rotation of embedding space perform poorly for Node2Vec embeddings even on the same graph and especially if they are used to align embedding on the different graphs, since differences between embeddings cannot be reduced to rotations. 

If generalization, either to the new graph or to unseen nodes on the same graph, is the goal, a common approach is to use message passing inductive models, most frequently implemented as graph neural networks \cite{GNN_Gori, GNN_Scarselli}. While such networks are tailored to aggregation of transformed features of graph nodes, they still can be used with featureless graphs \cite{on_node_features} using, among other approaches, 
random initialization \cite{surprising_power_random_initialization, sato2021random} or deterministic features (constant, degree of the node etc.) \cite{ node_features_for_gnn}

One recent paper \cite{node_features_for_gnn} investigates using centrality-based features for node classification using GNN. Although the authors focus on task specific training that does not aim to generalize between graphs, we have observed in our experiments that such features (with the exception of those that are not invariant over graph isomorphism, such as coloring number) can be used to build generalizable downstream models.

While we are not aware of any investigations comparing the relative performance of approaches to build task-agnostic, generalizable embeddings on unseen graphs, it is easy to see that centrality-based features, either by themselves or after GNN-type aggregation, can be generalizable for some tasks, since they are defined by the structure of the local neighborhood. For task-specific inductive models, the focus is usually on the task objective rather than on the node representation. However, such embeddings can be naturally constructed based on the outputs of different GNN layers. The task-agnostic property can be satisfied by training GNN for a predefined task, for instance, for link prediction or for some other structure-based objective, and using the resulting embeddings as input for the downstream task. 

Such techniques - the direct use of centrality-based features as well as embeddings generated by GNN based on those features and trained for link prediction - are used in this work as a baseline for comparison with our proposed approaches. Additionally, we investigate the benefits of combining these techniques with our proposed methods through ensembling. 

\subsection{Random projections}

Random projection for graph representation learning has been introduced in \cite{RandNE_paper} as the RandNE algorithm and in \cite{FastRP_paper} as FastRP. In RandNE, embeddings are created as 
\[U=\sum_{k=0}^{q}{\alpha_kA^kU_0}\]
where $U_0\in\mathbb{R}^{\left|V\right|\times D}$ is a random projection matrix which is created as a result of othogonalizatuon of Gaussian random matrix and A is the adjacency matrix of the graph. In \cite{RandNE_paper}, the authors used $q=3$ and employed grid search to find task specific coefficients $\alpha$.

In FastRP, embeddings are produced as
\[U=\sum_{k=1}^{q}{\alpha_kS^kLR}\]
where S is the transition matrix of the graph, $R\in\mathbb{R}^{\left|V\right|\times D}$ is the sparse random projection matrix 
\[R_{ij}=\begin{cases}\sqrt s\ with\ probability\ \frac{1}{2s} \\0\ with\ probability\ 1-\frac{1}{s}\\ -\sqrt s\ with\ probability\ \frac{1}{2s} \\ \end{cases}\]
and L is a normalization matrix
\[L=diag\left(\left(\frac{d_1}{2m}\right)^\beta,...,\left(\frac{d_{|V|}}{2m}\right)^\beta\right)\]
where $d_i$ is the degree of node and $m=|E|$ is the count of edges in the graph. In \cite{FastRP_paper}, the authors found that a maximum polynomial power $q=4$ was sufficient for the tasks that they explored. They also found that it was sufficient to have only two non-zero coefficients $\alpha_4$ and $\alpha_3$; one of them could be fixed to 1. In their experiments, the optimal value of $\beta$ was found to be $-0.9$.

Both algorithms use the fact that computation of $A^kR$ where $A$ is an adjacency or transition matrix of graph $G=(V,E)$ can be performed in $O(k|E|)$ time using chain multiplication 
\begin{equation} \tag{1}
A^k R\ = A (A^{k-1}R)
\end{equation}
since multiplication by such a matrix can be implemented as aggregation over each node's immediate neighborhood.

Using random projections as raw input to downstream models faces the same generalization problems as using vectors generated using other transductive approaches such as Node2Vec \cite{node2vec_paper}. On a single set of embeddings, models can achieve impressive performance, especially if they can rely on label propagation from training set to the neighboring nodes in the test set. However, such models will not work on unseen graphs since they will effectively try to compute distances between nodes in two disjoint graphs, which clearly cannot provide any useful signal. Except in artificial cases that place a lot of unrealistic assumptions on similarity between graphs or for very specific tasks (for example, computing distance to the root of a tree), label propagation models cannot be made generalizable by doing task-agnostic alignment between embedding spaces (for instance by running singular value decomposition). However, another means of inference can rely on retrieving isomorphic features from embeddings in a task agnostic way and passing them to a downstream model; that approach can be generalized. Models that use raw embeddings can operate in both modes. In this work, we are looking at ways to restrict the manner of model inference toward generalizable methods via either embedding preprocessing or by using specific a model architecture. We show that random projections provide a convenient and efficient way to do so.

\section{Our contribution}
Our goal is to create graph representations that are valuable for building pairwise and single-node models while remaining generalizable across different graphs.
We build on powers of the transition matrix of the graph to create these representations.
Given nodes $i,j$, consider dot products over rows of the powers of the transition matrix:
\[F_{i,j}^{(k,s)}={(A^k)}_{i,\ast}\cdot{(A^s)}_{j,\ast}\]
These dot products can be interpreted as the probability that a random walk from node $i$ of length $k$ and a random walk from node $j$ of length $s$ meet at the same endpoint.
Note that this probability is invariant over graph isomorphism -- a key ingredient for generalization across graphs.
By varying the lengths $k$ and $s$, we can collect a set of pairwise node features $F_{i,j}^{(k,s)}$.
Single node features can be computed by starting and ending random walk at the same node: $F_i^{(k,s)}\ =\ F_{i,i}^{(k,s)}$. Furthermore, we show that a neural model architecture taking advantage of random projection properties leads to results that outperform the direct estimation of features $F_{i,j}^{(k,s)}$ using dot products of random projections.

Below, we first describe the mathematical reasoning behind the use of random projections to approximate powers of transition matrices and then present two approaches for their use in practical applications.

\subsection{Random projections of powers of transition matrix}

For our approach we use random projections of matrices $A^k$ and $A^s$:

\begin{equation} \tag{2}  
\left(A^kR\right)_{i,*}\cdot\left(A^sR\right)_{j,*}={(A^kR{{(A}^sR)}^T)}_{i,\ j}\approx F_{i,j}^{\left(k,s\right)}
\end{equation} 

\noindent
where $A\in\mathbb{R}^{|V|\times|V|}$ is the transition matrix of the graph and $R\in\mathbb{R}^{|V|\times D}$ is a random matrix, to approximate  $F_{i,j}^{(k,s)}$ - dot products over rows of powers of the transition matrix $A$.

The intuition behind equation (2) can be derived from the following fact: if $R$ is a random matrix with i.i.d. elements drawn from some distribution $P$ with mean $E_{x~P(x)}(x)=0$, variance $E_{x~P(x)}(x^2)=\sigma^2$, and 4th central moment  $E_{x~P(x)}(x^4)=\mu_4$, then the matrix $RR^T$ has as its elementwise mean the identity matrix (multiplied by a scalar)
\begin{equation} \tag{3}
E(RR^T)\ =\ D\sigma^2I_{|V|}
\end{equation}
and elementwise variance for diagonal and non-diagonal elements 
\[\sigma_{diag}^2=D\ {(\mu}_4 - \sigma^4)=D(Kurt(P)-1)\sigma^4\]
\[\sigma_{nondiag}^2=D\ \sigma^4\]
For instance, using a gaussian initialization function $P=N(0,\ \frac{1}{D})$, the mean of the matrix $RR^T$ will be the identity matrix $I_{|V|}$ and the variance of its elements will be $\sigma_{diag}^2=\frac{2}{D}$ ,$ \sigma_{nondiag}^2=\frac{1}{D}$. Using (3) we can derive 
\[E_R(A^kR{{(A}^sR)}^T)={E_R(A}^kR{{(A}^sR)}^T)=A^k{{E_R(RR^T)(A}^s)}^T=A^k{{(A}^s)}^T\]
and 
\[E_R{(A^kR{{(A}^sR)}^T)}_{i,\ j}={(A^k{{(A}^s)}^T)}_{i,\ j}=F_{i,j}^{\left(k,s\right)}\]

Another way to derive (2) is based on Jonson-Lindenstauss lemma \cite{Johnson_Lindenstrauss_article}, the main result used for justifying random projections techniques, which states that for any $0<\varepsilon<1$ and a set $S$ of $m$ points in $\mathbb{R}^{|V|}$ there exists a linear map (which can be constructed using random projections  \cite{ElementaryProof_JL_Dasgupta}) to ${f:\ \mathbb{R}^{|V|}\rightarrow\mathbb{R}}^D$ where $D>\frac{8ln\ m}{\varepsilon^2}$  such that for any two points $u$ and $v$ from $S$
\[(1-\varepsilon)\lVert u-v \rVert^2 \leq \lVert f(u)-f(v)\rVert^2 \leq (1+\varepsilon)\Vert u-v \rVert^2\] 
This statement can be reformulated for dot products (if we add zero vector to the set $S$ and assume that $f(0)=0$, which is true for random projections)
\begin{multline}
|f(u)\cdot f(v)-\ u\cdot v|= \\ 
\frac{|(\lVert f(u)\rVert^2-\lVert u\rVert^2) + (\lVert f(v)\rVert^2-\lVert v\rVert^2)-(\lVert f(u) - f(v)\rVert^2-\lVert u-v\rVert^2)|}{2} \\ 
\leq 3\varepsilon max(\lVert u\rVert^2, \lVert v\rVert^2, \lVert u-v\rVert^2) \nonumber
\end{multline}
If vectors $u$ and $v$ represent the rows of some power of transition matrix, i.e. some probability distribution where the elements of the vector are positive and sum to 1, this inequality can be simplified as 
\[|f(u)\cdot f(v) - u\cdot v|\leq 6\varepsilon\] 

\subsection{RP DotProd and RP ConvNet}
Below we develop two methods for defining generalizable features based on these random projections.

The hypothesis that we will empirically evaluate states that if the latent process that creates the labels we are trying to predict uses, at least partially, graph structure to generate those labels, then the features $F^{(k,s)}$ will provide enough information about such structure to build reasonable models.

This hypothesis, together with using chain rule (1) for iteratively producing random projections of powers of transition matrix, leads to the following algorithm, that takes as an input the graph transition matrix $A$ and produces set of features $F$ for each node or pair of nodes to be used in the downstream model: 

\vspace{0.4em}
\noindent\rule{\columnwidth}{1.5pt}
\vspace{0.2em}\noindent{}\textit{\textbf{RP DotProd}: Dot product of random projections of powers of transition matrix}

\vspace{-0.5em}
\noindent\rule{\columnwidth}{0.75pt}
\noindent{}\textsc{input:} Graph transition matrix $A$

\noindent{}\textsc{output:} Features  $F_{i}^{(k,s)}$ for single node or $F_{i,j}^{(k,s)}$ for pair of nodes

\vspace{-0.5em}
\noindent\rule{\columnwidth}{0.25pt}
\begin{algorithmic}
\State Initialize random matrix $R^{(0)}\in\mathbb{R}^{|V|\times D}$ (gaussian, iid)
\For{$k$ in $[1..N]$}
     \State Compute and store $R^{(k)}\ =\ A\ R^{(k-1)}$
\EndFor
\State \textit{To compute single node features:}
\For{node $i \in [1..|V|]$}
\State \textit{compute set of $\frac{N(N+1)}{2}$ features}
\State $F_{i}^{(k,s)}=R_i^{(k)}\cdot R_i^{(s)}$ where $0 \le k \le N,\ k\le s \le N$
\EndFor

\State \textit{To compute pairwise features, compute dot products between projections of nodes:}
\For{nodes $i,j \in [1..|V|]$}
\State $F_{i,j}^{(k,s)}=R_i^{(k)}\cdot R_j^{(s)}$ where ${0\le k \le N, 0 \le s \le N}$.
\EndFor
\end{algorithmic}
\noindent\rule{\columnwidth}{1.5pt}
The resulting set  $\left\{F_{i,i}^{(k,s)},\ F_{i,j}^{(k,s)},F_{j,j}^{(k,s)}\right\}$ contains $(N+1)(2N+1)$ features.
These features become an input to downstream task model (for instance, a feed-forward neural network)

This approach isolates the downstream task from the raw values of embeddings $R^{(k)},\ k=0..N$. While this produces good relevance in both single-node and pairwise classification, empirically we have found that passing the embeddings themselves to downstream task can further improve relevance. However, allowing a very generic architecture of neural network to process these embeddings, i.e. without restricting the type of function it can learn, leads to difficulties in training generalization, since it will tend to overfit to the training set by learning to identify nodes/group of nodes in the training graph instead of extracting features from graph structure.

We had much better success in using a variant leveraging the fact that different slices $R_{\ast,p}^{(\ast)},\ p=1..D$ of random projections tensor are mutually independent but are drawn from the same distribution. Therefore, it makes sense to independently and uniformly process each dimension $p$ of the set of random projections of node or pair of nodes and to aggregate the resulting vectors using a set function (since order of dimensions is arbitrary) before passing them to the classifier. This architecture helps prevent the downstream model from learning artifacts of specific random projection initializations, without the need to train it on multiple versions of random projections to teach the model to only extract features that are independent of the initialization (random matrix $R$).

This leads to a second version of our approach, which takes as an input the graph transition matrix $A$ and produces the set $X$ of node features. Unlike the RP DotNet approach, which does not prescribe how features are used, here we recommend a specific architecture of the layer that transforms them before they are used in the rest of the model. The parameters of this transformation layer have to be jointly optimized with the rest of the task-specific model, which makes its output task-specific. Since our goal is to build embeddings that are task-independent, we use $X$ as the node representation and not the features ($F$) in the following method:

\vspace{0.4em}
\noindent\rule{\columnwidth}{1.5pt}
\vspace{0.2em}\noindent{}\textit{\textbf{RP ConvNet}: Processing of random projections of powers of transition matrix using task-specific neural network}

\vspace{-0.5em}
\noindent\rule{\columnwidth}{0.75pt}
\noindent{}\textsc{input:} Graph transition matrix $A$ 

\noindent{}\textsc{output:} Feature vector $F_{i}$ for single node or $F_{i,j}$ for pair of nodes

\vspace{-0.5em}
\noindent\rule{\columnwidth}{0.75pt}
\begin{algorithmic}
\State Initialize random matrix $R^{(0)}\in\mathbb{R}^{|V|\times D}$ (gaussian, iid)
\For{$k$ in $[1..N]$}
    \State Compute and store $R^{(k)}\ =\ A\ R^{(k-1)}$
\EndFor
\State \textit{Build representation for each node $i=1..|V|$ as matrix $X^{(i)}\in\mathbb{R}^{D\times(N+1)}$:}
\[X_{p,k}^{(i)}=R_{i,p}^{(k-1)},\ p=1..D,\ k=1\ ..N+1\]
\State\textbf{These representations, by construction, are task-agnostic, while the rest of the model is optimized for the particular task together with the downstream model.}
\State \textit{For single node features, use representation $X^{(i)}$ directly }
\[Z^{(i)}=X^{(i)}\]
\State \textit{To compute pairwise node features, for each pair of nodes in the training or test set $i\in 1..|V|$, $j\in 1..|V|$:}
\State \indent Build the matrix  $Z^{(i,j)}\in\mathbb{R}^{D\times2(N+1)}$ by concatenating rows of $X^{(i)}$ and $X^{(j)}$
\[Z^{(i,j)}_{p,k}=\begin{cases}X^{(i)}_{p,k}, k \leq N+1 \\X^{(j)}_{p,k-N-1}, k>N+1\end{cases}, p=1..D,k=1 ..2(N+1)\]
\State \textit{Apply a transform $g_{\Psi}:\mathbb{R}^{N+1}\rightarrow\mathbb{R}^M$ ($M$ is a hyperparameter of the model) or $g_{\Psi}:\mathbb{R}^{2(N+1)}\rightarrow\mathbb{R}^M$ (for pairwise features) to the rows of the matrix $Z$ (one common implementation is to apply one or more layers of one-dimensional CNN with a non-linear activation function in between) to build the matrix $H\in\mathbb{R}^{D\times M}$ for each node }
\[H_{p,\ast}^{(i)}=\ g_{\Psi}(Z_{p,\ast}^{(i)}),\ \ p=1..D\]
\textit{or pair of nodes} 
\[H_{p,\ast}^{(i,j)}=\ g_{\Psi}(Z_{p,\ast}^{(i,j)}),\ \ p=1..D\]
\State \textit {Aggregate rows of matrix $H$ to produce vector $F^{(i)}\in\mathbb{R}^M$ or $F^{(i,j)}\in\mathbb{R}^M$ by computing the mean} 
\[F_{i}=\frac{1}{D}\sum_{p=1}^{D}H_{p,\ast}^{(i)}\]  \textit{or}  \[F_{i,j}=\frac{1}{D}\sum_{p=1}^{D}H_{p,\ast}^{(i,j)}\]
\end{algorithmic}
\noindent\rule{\columnwidth}{1.5pt}
\vspace{0.5em}

Vector $F \in \mathbb{R}^M$ is the feature vector for a node or pair of nodes that will be used by the downstream model to make the final prediction. It is assumed that we can compute the gradient of the loss function with respect to this vector, which will allow us to optimize the parameters $\Psi$ of the transform $g_{\Psi}$ (for instance, the weights of a 1-dimensional CNN). In practice, this usually means that the downstream model is a differentiable neural network itself, and we are combining both models into a single model that is optimized by a machine learning framework such as PyTorch or TensorFlow.

We used the mean to compute $F$ from the rows of $H$, but it is possible to experiment with a variety of aggregation operators developed for graph neural networks.

We compare our approaches to other techniques such as rotation-invariant features based on Node2Vec \cite{node2vec_paper}, centrality-based graph features \cite{node_features_for_gnn}, and embeddings produced by GNN trained on link prediction objective. In the case of pairwise prediction with Node2Vec, we also demonstrate how to create more features based on Node2Vec embeddings, which leads to a significant boost in the accuracy of downstream models: we can incorporate both input and output vectors for the two input nodes, computing cross-dot products of all available vectors. 

To run RP DotProd or RP ConvNet, one has to precompute random projections for all nodes in each graph the models will be applied to. The computational complexity of this operation is $O(|E|)$, which provides substantial benefits compared to the exact computation of powers of the transition matrix, which has complexity $O(|V|*|E|)$. At inference time, the models only need to use the features of the relevant node or pair of nodes, so the inference run time does not depend on the graph structure. 

Finally, it should be noted that it is easy to construct examples where any generalizable approach, including ours, will not work. One simple example would be a process that assigns a label to each node based on its proximity to a specific node $n_0$ in the graph. Transductive embedding methods such as Node2Vec will work to some extent in this example on the same graph. However, since the classifier will have to implicitly encode the identity of $n_0$, it will not work on any other graph where embeddings for $n_0$ have no meaning.

\section{Comparison approaches}
To measure the effectiveness of these two approaches in task relevance and generalization, we also evaluate several alternative methods for deriving node representations from a graph.

\vspace{0.3em}
\noindent{}\textit{\textbf{RI-Node2Vec}: Rotation invariant features based on Node2Vec embeddings}

Node2Vec \cite{node2vec_paper} and DeepWalk \cite{deepwalk_paper} models try to approximate the probability that node $j$ will appear in randomly chosen position of random walk with some predefined length starting from node $i$ as:
\begin{equation} \tag{4}
p_{ij}=\frac{e^{x_iy_j}}{\sum_{k}\ e^{x_iy_k}}     
\end{equation}
where $x$ and $y$ are vectors of same dimensionality $D$. The need to create two vectors for each node arises from the fact that $p_{ij}$ is asymmetric.

So as an input for single node inference model we can use two vectors $(x_i, y_i)$  and for pairwise inference - set of 4 vectors $(x_i,\ y_i,\ x_j,\ y_j)$. It is easy to see that (4) is invariant with respect to rotation transform, so any feature that can be used for building  generalizable model should at least be rotation invariant. If we consider the set of embeddings for all nodes, one can use different techniques such as singular value decomposition that can convert embeddings into some canonical form. We use a simpler
setting in which we are looking at extracting rotation invariant features from just the vectors for the nodes $i$ and $j$. In general, for set of vectors  ${v_p\in\mathbb{R}^D,\ p=1..K}$ any rotation invariant feature that is based solely on this set can be expressed as function of dot products of such vectors
\[F({v_p\in\mathbb{R}^D,\ p=1..K}\ )=\ F({v_i\cdot\ v_j,\ i=1..k,\ j=i..k})\]
this can be derived from following the Gram-Schmidt process and noticing that it defines the rotation against which $F$ should be invariant by construction. Therefore we can limit the features we are passing to the downstream model to those dot products.

\vspace{0.3em}
\noindent{}\textit{\textbf{IGF}: Invariant graph-based features}

Following \cite{node_features_for_gnn} we are using a set of graph isomorphic invariant features to provide some information about the local structure. Such features are
\begin{itemize}
\item Node degree
\item Node PageRank
\item Number of triangles node participates in 
\item K-core number
\item Largest clique number
\item Number of edges in egonet
\item Number of edges connecting egonet to the rest of the graph  
\end{itemize}
	
\vspace{0.3em}
\noindent{}\textit{\textbf{LinkGNN}: Link prediction embeddings generated by graph neural network}

Another way to produce embeddings for nodes in the graph is to run an inductive model to map nodes into some vector space $\mathbb{R}^D$. Such a model has to be trained for a particular objective. For the purposes of evaluation, we use the following experiment settings
\begin{itemize}
\item Use graph based features (the same as above) as input
\item Apply 2-layer GNN (we have used graph convolution network \cite{semisupervised_class_gnn}) to create node embeddings
\item Concatenate embeddings for the nodes in the pair and use 2 fully connected layers to predict if nodes are connected in the graph (we use graph edges as positive examples and random pairs of unconnected nodes as negative examples)
\item Train GNN to minimize cross-entropy loss of such prediction
\item After training, use GNN to produce embeddings for all the nodes
\end{itemize}

\section{Evaluation}

\subsection{Business Category in Yelp dataset}

For the evaluation of the generalization properties of different models, we first present experiments on the Yelp Open Dataset \cite{YelpDataset}. An attractive property of this dataset is that it is naturally segmented into disjoint graphs sharing similar statistical properties. 

This dataset consists of records of 150,346 businesses and 6,990,279 reviews for those businesses by 1,987,929 unique users. The businesses are selected from 11 different metropolitan areas, but those areas are not explicitly specified, and users can review businesses in different areas. Since the business records have latitude/longitude coordinates, we clustered them by distance into 11 clusters and used the cluster id to assign businesses to one of 11 areas (in the tables below, we use the state/province of the cluster centroid to identify the area, but there are some areas that span multiple states). After that, we formed 11 separate graphs containing businesses from a given area and users who reviewed them. (If a user reviews a business in two or more different areas, there will be multiple nodes representing that user in different graphs, but there will be no connection between those nodes.)

Businesses can be assigned to multiple categories simultaneously, but some of them are subcategories of another (for instance ``Doctors, Traditional Chinese Medicine''). We kept only the top-level category - there are 21 of them. The restaurant category remains the most frequent for all areas if taken individually (34\% of all businesses). Assigning the category to Restaurant provides a simple baseline performance to which we will compare the performance of different approaches. In each case, the training data consists of data for all areas except the one being evaluated -- effectively a type of cross-fold validation. For random projection approaches, the model is trained over 10 epochs, and model selection is done based on performance on the validation set consisting of samples taken from all other areas. Therefore, the test area is never used for either model training or model selection.

In Table~\ref{tblCatPred}, we compare baseline performance to two approaches that do not involve random projection as well as two approaches that use random projections. Both approaches that do not use random projections are utilizing LightGBM classifier \cite{ke2017lightgbm} with either graph features or embeddings from GNN that aggregates such features for link prediction (in this case, presence of review between user and business). Link prediction GNN is shared between and trained on data from all areas.

Random projection models use 128-dimensional random projection vectors to approximate the powers up to 10 of the business-to-business transition matrix. The matrix is not very sparse on its own; however, since the graph is bipartite, the business-to-business transition can be computed based on the sparse business-to-user and user-to-business transition matrices.
 
RP DotProd computes the dot product of vectors representing different powers of the transition matrix and passes them to a feedforward network with two 128-dimensional hidden layers. 
RP ConvNet uses a 1-dimensional CNN with 2 layers with 64 channels, and employs a feedforward network with a single 64-dimensional hidden layer for classification. 

\begin{table*}[h!]
\begin{tabular}{lrrrrrrr}
\toprule
Area &
\mcrot{1}{l}{60}{\parbox {1.8cm} {Number of\\ businesses}} &
\mcrot{1}{l}{60}{Baseline} &
\mcrot{1}{l}{60}{IGF} &
\mcrot{1}{l}{60}{LinkGNN} &
\mcrot{1}{l}{60}{RP DotProd} &
\mcrot{1}{l}{60}{\parbox {1.8cm} {RP ConvNet\\2x64, ReLU}} &
\mcrot{1}{l}{60}{\parbox {1.8cm} {RP DotProd \\ + IGF}} \\
\midrule
AB & 5573 & 0.432 & 0.224 & 0.319 & 0.425 & \textbf{0.466} & 0.437 \\
AZ & 9916 & 0.270 & 0.325 & 0.318 & 0.310 & \textbf{0.339} & \textbf{\textit{0.342}} \\
CA & 5177 & 0.224 & 0.262 & 0.252 & 0.272 & \textbf{0.306} & 0.301 \\
FL & 26336 & 0.332 & 0.361 & 0.355 & 0.366 & \textbf{0.376} & \textbf{\textit{0.386}} \\
ID & 4471 & 0.291 & 0.319 & 0.318 & 0.312 & \textbf{0.329} & \textbf{\textit{0.341}} \\
IN & 11248 & 0.369 & 0.382 & \textbf{0.397} & 0.386 & 0.386 & \textbf{\textit{0.409}} \\
LA & 9925 & 0.367 & 0.385 & 0.364 & \textbf{0.413} & 0.389 & \textbf{\textit{0.421}} \\
MO & 13062 & 0.401 & \textbf{0.403} & 0.392 & 0.392 & 0.397 & \textbf{\textit{0.405}} \\
NV & 7732 & 0.217 & 0.278 & 0.272 & 0.255 &\textbf{0.302} & 0.264 \\
PA & 44845 & 0.378 & 0.389 & 0.375 & 0.395 &\textbf{0.401} & 0.397 \\
TN & 12058 & 0.361 & 0.376 & 0.357 & 0.386 &\textbf{0.391} & \textbf{\textit{0.406}} \\
\midrule
Mean (equal weight) & & 0.331 & 0.337 & 0.338 & 0.356 &\textbf{0.371} & \textbf{\textit{0.374}} \\
Mean (weighted by number of businesses) &  & 0.348 & 0.361 & 0.356 & 0.371 &\textbf{0.381} & \textbf{\textit{0.385}} \\
\bottomrule
\end{tabular}
\caption{Category prediction accuracy for several model families on Yelp dataset.}
\label{tblCatPred}
\end{table*}

\begin{table*}[h!]
\begin{tabular}{lrrrcrrrr}
\toprule
& \multicolumn{3}{c}{Task GNN} && \multicolumn{4}{c}{RP ConvNet vs} \\ \cmidrule{2-4} \cmidrule{6-9}
Area &
others &
all  &
gain (all-others) &
&
IGF &
LinkGNN &
TaskGNN (others) &
RP DotProd \\
\midrule
AB & 0.231 & 0.478 & 0.247 && 0.242 & 0.147 & 0.235 & 0.041 \\
CA & 0.245 & 0.347 & 0.102 && 0.044 & 0.055 & 0.061 & 0.035 \\
ID & 0.345 & 0.398 & 0.054 && 0.010 & 0.012 & -0.015 & 0.017 \\
TN & 0.363 & 0.388 & 0.025 && 0.015 & 0.035 & 0.029 & 0.006 \\
NV & 0.283 & 0.299 & 0.016 && 0.024 & 0.029 & 0.019 & 0.047 \\
LA & 0.381 & 0.395 & 0.014 && 0.004 & 0.025 & 0.008 & -0.024 \\
AZ & 0.340 & 0.353 & 0.012 && 0.014 & 0.021 & -0.001 & 0.028 \\
PA & 0.387 & 0.397 & 0.010 && 0.012 & 0.025 & 0.014 & 0.006 \\
MO & 0.406 & 0.413 & 0.007 && -0.006 & 0.005 & -0.010 & 0.004 \\
IN & 0.402 & 0.408 & 0.006 && 0.005 & -0.011 & -0.016 & 0.001 \\
FL & 0.368 & 0.373 & 0.005 && 0.014 & 0.020 & 0.007 & 0.010 \\
\midrule
Correlation & - & - & \textbf{1.000} && \textbf{0.957} & \textbf{0.940} & \textbf{0.952} & \textbf{0.527} \\
\bottomrule
\end{tabular}
\caption{RP-CNN gain vs comparison and relevance gain from adding test graph into train (as measure of distribution deviation).}
\label{tblCatPredGraph}
\end{table*}

The metric that is used is the accuracy of class detection. Random projection techniques outperform other approaches on average as well as individually for 9 out of 11 areas. The convolutional model is better than the dot product model in 10 out of 11 areas. An interesting case is Alberta (AB), where techniques based on graph features significantly underperform compared even to a simple baseline, though the random projections methods seem to generalize to that area in a much more robust manner.

We also investigated hybrid approaches in which graph features and GNN embeddings are added to the input of the feedforward network of the random projection models. The rightmost column in the table represents the best-performing technique that combines the dot products of random projections with graph features. For 10 out of 11 areas, it improves on both the methods it ensembles and is the best-performing technique on average and for 7 out of 11 areas taken individually. 

Finally, we investigated the hypothesis that the suggested approaches behave more robustly in situations where the statistics of the training graph deviate from the training data, i.e., that they capture more generalizable features. To gauge this deviation, we measure the accuracy gain from training on all areas versus excluding the test area if we are training the task-specific GNN model.  which directly predicts the category (we use a 2-layer GCN (from \cite{semisupervised_class_gnn}) with a hidden dimension size of 128, which takes as input the same graph features as the non-task-specific models above and uses a 2-layer FCN with the same hidden dimension size as the decision layer). We compute the mean test set accuracy on the last 10 epochs of the 100 epochs of training performed on all graphs ("all") and on all graphs excluding the test area ("others") to reduce the volatility of prediction in the "others" case. 

Table~\ref{tblCatPredGraph} shows the relationship between this measure (accuracy gain from adding test graph data to the task-specific GNN training set) and the improvement in accuracy of the RP ConvNet model versus alternative approaches, including the task-specific GNN (trained on data that does not include the test area), as well as the correlation coefficient between those values. Results seem to be consistent with the hypothesis and show that, at least for this dataset, the RP ConvNet model is more robust against changes in graph statistics.
\subsection{Amazon Computers and Amazon Photo network}
For additional evaluation, we have used the dataset introduced in \cite{pitfalls_gnn_eval}, which contains nodes representing goods and edges between nodes representing the fact that those goods are frequently purchased together. The goods are assigned to categories (node classes). In the original dataset, nodes also have features (encoded product reviews), but we do not use these features in our evaluation.  The Computers dataset has 13,381 nodes belonging to 10 classes, connected by 245,778 edges. The Photo dataset has 7,487 nodes in 8 classes, connected by 119,943 edges.

Evaluation is done on two tasks: node classification (product category detection) for single-node inference and binary classification of pairs of nodes (regardless of whether they have an edge between them in the graph) to predict if such nodes have the same category. In the second task, the actual labels of the nodes are not available for the training process.

In our experiments, the Computers dataset has been split into two parts (by randomly assigning 50\% of the nodes to each of them) and edges connecting nodes in different parts have been discarded. As a result, we have two disjoint graphs that are known to be drawn from the exact same distribution with respect to both graph connectivity and label assignment. This assumption is usually unrealistic if the graph represents different entities, such as different organizations. However, it allows us to evaluate generalization in ideal conditions, with the understanding that in real-world settings, performance will be worse. 
We also ran evaluations on the Photo dataset, which was only used for testing models trained on the split of the Computers dataset. While the pairwise task of predicting whether two nodes are in the same category can be directly evaluated (since it does not depend on actual labels), single node classification is more difficult as the categories of nodes are different between the two datasets. Instead, for the Photo dataset we compute accuracy by assuming the optimal mapping between predicted and actual class labels. This metric can be computed by grouping nodes into clusters based on predicted labels and assigning the mode of the actual labels to the whole cluster.

To train the downstream task, we used a standard train/test split of nodes in the train part of the Computers dataset (reserving 33\% of nodes for the test set). In all cases, model selection is performed based on the loss on the test split of the train graph. For cases where we have performed some hyperparameter selection (e.g. number of layers or hidden dimensions), we only show the model with the best performance on the test graph derived from the Computers dataset.


Metrics are shown for all four sets: train and test splits of the train graph, test split of the Computers dataset, and the Photo dataset. For single node classification, we report the accuracy of prediction and for the pairwise classifier, we report the area under the ROC curve (AUC).

Performance metrics on the last two sets are, in a somewhat different way, indicative of the generalization capabilities of the models. The performance on the train split of the train graph (1st set) is only given to illustrate the model's capacity, and models with larger feature vectors unsurprisingly demonstrate better results on that split. High performance on the test split of the train graph (2nd set) is harder to obtain but still correlates with model capacity. However, as demonstrated by the GNN embeddings, which perform best on that set, outperforming other models on the same graph does not necessarily lead to superior performance on a different graph, even one sharing the same statistical properties (i.e. the test split of the Computers dataset). The most likely explanation for this behavior is that the large size of the embedding vector, even one produced by an inductive model, allows the downstream model to pick up features that are useful for making predictions on the same graph, but do not translate to another one.

Performance on the Photo dataset is shown as an illustration of the generalization capabilities of the model to a graph that is related, but has different statistical properties. However, it can be difficult to draw conclusions based on that metric in this task because it is not clear to what extent the performance on that dataset can be attributed to better generalization capabilities versus luck in focusing on features that happen to be more transferrable between the two datasets. A proper investigation of generalization across different datasets picked from the same distribution requires the ability to use multiple such datasets for both training and testing (as was done in the case of the Yelp dataset in the previous section).

In all cases, the downstream model was implemented as a fully connected network with ReLU activation. The hidden dimension sizes and number of layers are listed in the tables. For RP ConvNet, we used a 2-layer CNN with a hidden dimension size of 64.

\textbf{Results for classification task}

\begin{table*}
\begin{tabular}{lrrrrrrr}
\toprule
Area &
\mcrot{1}{l}{60}{Baseline} &
\mcrot{1}{l}{60}{RI-Node2Vec} &
\mcrot{1}{l}{60}{IGF} &
\mcrot{1}{l}{60}{LinkGNN} &
\mcrot{1}{l}{60}{RP DotProd} &
\mcrot{1}{l}{60}{\parbox {2cm} {RP ConvNet \\ 2x64, ReLU}} &
\mcrot{1}{l}{60}{\parbox {2cm} {RP ConvNet \\ + IGF}} \\
\midrule
Computers, Train Graph, Classifier Test & 0.374 & 0.431 & 0.524 & \textbf{0.963} & 0.587 & 0.642 & 0.646 \\
Computers, Train Graph, Classifier Train & 0.372 & 0.427 & 0.467 & \textbf{0.703} & 0.509 & 0.579 & 0.587 \\
Computers, Test Graph & 0.377 & 0.437 & 0.440 & 0.472 & 0.476 & \textbf{0.526} & 0.523 \\
Photo & 0.256 & 0.339 & 0.360 & \textbf{0.437} & 0.361 & 0.368 & 0.424 \\
\midrule
FCN layers/hidden size& & 2x128 & 2x128 & 2x128 & 2x128 & 1x64 & 1x64 \\
\bottomrule
\end{tabular}
\caption{Category prediction accuracy on Amazon dataset}
\label{tblAmazonClassification}
\end{table*}

In Table~\ref{tblAmazonClassification}, we show results for single node classification models as well as a baseline performance, which is just prediction of the most common outcome regardless of input. Random projection techniques (RP DotProd and RP ConvNet) outperform the best alternative (embeddings from the GNN trained on the link prediction task) on the test split of the Computers dataset. At the same time, they do not perform as well on the Photo dataset, which, as discussed above, does not necessarily indicate that they will generalize worse if trained on multiple datasets. The best performing ensembling technique, which combines RP ConvNet with graph features, is a close second to the best single model approach on both datasets (RP ConvNet on the Computers test and GNN embeddings on the Photo dataset).

\textbf {Results for pairwise prediction (same class detection)}

\begin{table*}
\begin{tabular}{lrrrrrrr}
\toprule
Area &
\mcrot{1}{l}{60}{\parbox {3.5cm} {RI-Node2Vec \\Output only - 3 features}} &
\mcrot{1}{l}{60}{\parbox {3.5cm} {RI-Node2Vec \\Input/Output - 10 features}} &
\mcrot{1}{l}{60}{IGF} &
\mcrot{1}{l}{60}{LinkGNN} &
\mcrot{1}{l}{60}{RP DotProd} &
\mcrot{1}{l}{60}{\parbox {3.5cm} {RP ConvNet \\2x64, ReLU}} &
\mcrot{1}{l}{60}{RP ConvNet + IGF} \\
\midrule
Computers, Train Graph, Classifier Test & 0.600 & 0.833 &  0.720 & \textbf{0.893} & 0.828 & 0.866 & 0.860 \\
Computers, Train Graph, Classifier Train & 0.609 & 0.831 & 0.718 & \textbf{0.865} & 0.817 & 0.844 & 0.851 \\
Computers, Test Graph & 0.605 &0.821 & 0.703 & 0.705 & 0.815 & \textbf{0.838} & 0.828 \\
Photo & 0.605 & \textbf{0.791} & 0.540 & 0.636 & 0.746 & 0.754 & 0.728 \\
\midrule
FCN layers/hidden size& 2x512 & 2x512 & 2x512 & 2x512 & 2x512 & 2x64 & 2x64 \\
\bottomrule
\end{tabular}
\caption{Same class (pairwise) prediction AUC on Amazon dataset, baseline is 0.5 }
\label{tblAmazonPairwise}
\end{table*}

Results for pairwise classification are listed in Table~\ref{tblAmazonPairwise}. For this task, we are comparing two approaches for using Node2Vec embeddings. In one, we are considering only the input vectors of Node2Vec, which for two nodes creates 3 features (squared norms and dot product of vectors). In the other, features are built using cross dot products of all 4 vectors (input and output vectors for each node) – a total of 10 features. The model that uses 3 features is the weakest of all the approaches we have investigated, but the second one, using all 4 vectors, has much better performance, outperforming all other models that do not use random projections. We are not aware of such an approach (using cross dot products of 4 Node2Vec vectors for pairwise prediction) being explicitly described in the literature. Therefore, we would like to highlight that it can improve performance over norms and dot product models, or even more over the single distance feature (usually dot product, cosine similarity, or Euclidian distance) in many scenarios where Node2Vec embeddings need to be used in a generalizable way, i.e. without supplying raw embeddings to the downstream model.

The RP ConvNet approach is marginally outperforming Node2Vec with 4 vectors on the test graph (but not on the Photo dataset). It is possible that such gain can be reversed by tuning hyperparameters, such as the parameters of the Node2Vec random walk or the architecture of the downstream model.

However, random projections are generally much cheaper to compute than Node2Vec vectors, which require multiple steps of gradient descent for their computation. Therefore, if computational costs are a significant decision factor, random projection can be preferred to Node2Vec based on cheaper computation, given that they have very similar performance.

In Table~\ref{tblAmazonPairwise}, we have also listed the results for the best performing ensembling model. However, it does not seem that ensembling gives any benefits in this case.

\section{Conclusion}

We have suggested approaches for building task-agnostic node representations of featureless graphs that allow us to build single-node and pairwise task-specific models that are generalizable, i.e., can be applied to graphs that were unseen during training. The representations are produced using computationally efficient propagation of random messages through the graph in a way that is similar to FastRP \cite{FastRP_paper} and RandNE \cite{RandNE_paper}, but unlike those techniques, we do not use an aggregation bottleneck and represent the node by a set of random projection vectors. This allows us to extract multiple features from a single node and even more from a pair of nodes. It can be shown that such features approximate isomorphic functions on the graph (the probability of random walks meeting at the same point) and therefore, models that use them are generalizable in the sense that, if we have a large enough training and test set of graphs drawn from the same population, the performance on the test and training set will converge. We also suggest a specific architecture of a neural network that does not explicitly build such features but uses the properties of random projection to achieve the same generalization objective, outperforming dot product features in our experiments.

We conducted an empirical analysis that compares the suggested approaches to other techniques that also lead to the creation of generalizable models. We are not aware of any evaluations in the same settings (evaluation of a model that consumes task-agnostic node representations and is done on an unseen graph), so we could not rely on published results for other approaches and had to run evaluations ourselves. In our experiments, the suggested approaches appear to outperform the best alternatives in many cases and closely match them in the rest of the situations, but we are not making the claim that the suggested techniques outperform other approaches in the general case - in fact, it is easy to construct a dataset that will favor any given approach. However, given the computational efficiency of random projections, our results suggest that in many practical tasks, it would make sense to consider them if the task being solved matches our settings. Additionally, in some of our experiments, building an ensemble of random projections with other techniques gives a performance boost over either approach taken separately, which can be of value in some scenarios. 

Finally, although we did not run the evaluation for such tasks, the random projection approach can be naturally extended to directed graphs and/or graphs with positively weighted edges. Another possible research area is the application of this technique to heterogeneous graphs. In the latter case, we can associate the node with not one but a set of matrices representing different metapaths.

\newpage
\bibliographystyle{acm}
\bibliography{reflist_rp}
\end{document}